\documentclass[conference]{IEEEtran}
\usepackage{cite}
\usepackage{amsmath,amssymb,amsfonts}
\usepackage{algorithmic}
\usepackage{graphicx}
\usepackage{textcomp}
\usepackage{xcolor}
\usepackage{hyperref}
\usepackage{listings}
\usepackage{makecell}

\def\BibTeX{{\rm B\kern-.05em{\sc i\kern-.025em b}\kern-.08em
    T\kern-.1667em\lower.7ex\hbox{E}\kern-.125emX}}

\newcommand{\algo}{\textsc{FairLabel}}

\begin{document}

\title{\algo: Correcting Bias in Labels}

\author{\IEEEauthorblockN{1\textsuperscript{st} Srinivasan H Sengamedu}
\IEEEauthorblockA{\textit{People Experience and Technology} \\
\textit{Amazon}\\
Seattle, USA \\
sengamed@amazon.com}
\and
\IEEEauthorblockN{2\textsuperscript{nd} Hien Pham}
\IEEEauthorblockA{\textit{People Experience and Technology} \\
\textit{Amazon}\\
Seattle, USA \\
hienpham@amazon.com}
}

\maketitle

\begin{abstract}
There are several algorithms for measuring fairness of \textit{ML models}. A fundamental assumption in these approaches is that the ground truth is fair or unbiased. In real-world datasets, however, the ground truth often contains data that is a result of historical and societal biases and discrimination. Models trained on these datasets will inherit and propagate the biases to the model outputs. We propose \algo, an algorithm which detects and corrects biases in labels. The goal of \algo is to reduce the Disparate Impact (DI) across groups while maintaining high accuracy in predictions. We propose metrics to measure the quality of bias correction and validate \algo\ on synthetic datasets and show that the label correction is correct 86.7\% of the time vs. 71.9\% for a baseline model. We also apply \algo\ on benchmark datasets such as UCI Adult, German Credit Risk, and Compas datasets and show that the Disparate Impact Ratio increases by as much as 54.2\%.
\end{abstract}

\IEEEpeerreviewmaketitle

\section{Introduction}
With ML models playing a fundamental role in many decisions such as job applications, loan applications, and criminal justice decisions, Algorithmic Fairness has emerged as an important aspect of ML modeling. Typically, societal bias and historical discrimination manifest both in adverse decisions against the minority group and favorable decisions for the majority group. In this paper, we refer to the majority group as receiving favorable treatment and the minority group as receiving unfavorable treatment. The biases in decisions, whether conscious or unconscious, lead to biased data. ML models trained on these data, if steps are not taken to mitigate the inherent bias contained within them, can lead to biased model outputs and decisions. This leads to further propagation of bias. 

In Algorithmic Fairness, there are several metrics such as \textit{Disparate Impact Ratio (DIR)} that are defined to quantify the fairness of ML models. The U.S. Equal Employment Opportunity Commission has established a guideline termed the \textit{Four-Fifth's or 80\% Rule} which states that the selection rate for the minority class should not be less than four-fifths (80\%) of that of the majority group. See~\cite{usoc}~and~\cite{nyc}. Real world datasets such as UCI Adult dataset has a DIR ratio of male/female (salaries above \$50K USD per annum) of 0.353 which is well below the acceptable threshold of 0.8 for disparate impact.

Manually correcting ground truth labels in real-world datasets is an impossible task as this means re-assessing historical decisions (such as loan and job applications) for which detailed data and information are no longer available. Instead, we propose an algorithm, \algo, to correct biased labels directly. We propose a data generation framework to validate \algo in which we inject bias into the synthetic data and measure the algorithm's ability to find and correct the bias. We then, we show the performance of \algo\ on ML benchmark datasets such as UCI Adult, German Credit Risk, and Compas.

The contributions of the paper are as follows.
\begin{enumerate}
    \item We propose \algo, an algorithm to identify and correct biases in labelled data.
    \item We propose a framework to generate biased synthetic data to validate \algo. We also propose relevant metrics for the task.
    \item We demonstrate the performance of \algo\ on UCI Adult, German Credit Risk, and Compas datasets and report improvements in DIR. The improvements range from 13.4\% to 54.2\%.
\end{enumerate}

The paper is organized as follows. Section~\ref{relwork:sec} discusses related work in the area of algorithmic fairness. Section~\ref{algo:sec} describes \algo\ algorithm. Section~\ref{syngen:sec} outlines the synthetic data generation framework which introduces noise and bias to mimic real-world data and Section~\ref{metrics:sec} defines relevant metrics for the debiasing task. Section~\ref{expt:sec} contains details of synthetic and real-world datasets used and the performance of \algo\ in terms of the debiasing metrics as well as Disparate Impact. Section~\ref{conc:sec} concludes the paper.

\section{Related Work} \label{relwork:sec}
There is rich research work on Algorithmic Fairness. A recent review on Bias and Fairness~\cite{bias-survey-acm-2021} lists the sources of bias as well as techniques to improve fairness. The sources of bias are often listed as biased features, selection bias, and label imbalance. Algorithmic bias is defined as \textit{bias is not present in the data and is added purely by the algorithm}. Hence the problem of label bias has not received been addressed widely. For another, more recent survey, see~\cite{bias-survey-acm-2022}. The book~\cite{barocas-hardt-narayanan} provides a broad as well as deep coverage of the topic.

The standard framework for achieving algorithmic fairness is either using pre-processing, mid-processing, or post-processing to achieve target fairness metrics. As mentioned earlier, all these frameworks \textit{assume that the data is unbiased} and the unfairness springs mainly from feature representations or ML models.
\cite{6175897} discusses  prevention of discrimination in data mining. Pre-processing approaches include messaging \cite{4909197}, preferential sampling \cite{Kamiran2012}~\cite{kamiran2010classification}, disparate impact removal \cite{10.1145/2783258.2783311} to remove biases from the data. Papers like~\cite{10.1145/2783258.2783311} change features but \textit{keep labels intact} while removing data biases. \algo\ is complementary in the sense that it removes label bias.

There is relatively less research focus on biases in data. Biases in data often means selection or curation bias and biases in features in terms of missing features or missing values.
One challenge in data collection is data skews~\cite{gebrudatasheets}~\cite{benjamintowards}~\cite{bender-friedman-2018-data}~\cite{Mitchell:2019:MCM:3287560.3287596}. Analysis often happens on slices of data and patterns in individual slices and aggregates can be different. This is called Simpson Paradox~\cite{simpson}.
\cite{kievit2013simpson}~\cite{alipourfard2018wsdm}~\cite{alipourfard2018icwsm} address Simpson Paradox related issues. \cite{holland2018dataset}  propose having labels to categorize data quality. Previous work has leveraged different loss function~\cite{jiang2020identifying} instead of correcting data bias.

Real-world data often has biased labels. Getting the data relabeled is usually not an option.  At the same time, it is essential to use historical data to build models. However, the issue of bias in ground truth labels and debiasing them has received relatively less research focus. Synthetic data generation has been used to mitigate the problem~\cite{xu2023ffpdg}~\cite{Madl2023ApproximateAA}. However, these methods care more about privacy than bias. 

Counterfactual Analysis is a popular causal analysis approach. Research on Counterfactual Fairness~\cite{cft-fair} mentions the issues of biased ground truth. The approach taken in Counterfactual Fairness is changing \textit{feature values} to achieve fairness. The approach is based on structural models with latent variables. The challenge with this approach is the is the validation. \cite{zhang2017achieving} also uses causal approaches. 

FairSMOTE~\cite{fair-smote} is an algorithm to detect biased labels. FairSMOTE is based on 'situation testing': flip the sensitive attribute and check if the label has changed. The paper does not propose metrics to define the effectiveness of the approach. It is interesting to note that only the training data is debiased and not the test data. The reason is that the model solving the task is also used for debiasing. We decouple the two tasks and use a separate model for label debiasing. 

The challenge with these approaches is that they do not consider the fact that \textit{bias is asymmetric or directional and unbalanced.} In other words, one demography is penalized and the other demography is advantaged (\textit{asymmetry}) and one of the demography is often under-represented (\textit{unbalanced}). We need principled approaches to deal with these issues.

Appendix~\ref{metrics:app} lists metrics used in Algorithmic Fairness. We primarily use Disparate Index Ratio from an application perspective. For the newly defined task of label debiasing, we define metrics in Section~\ref{metrics:sec}.

\section{\algo} \label{algo:sec}
The intuition behind \algo\ is simple. In the minority class, a biased decision occurs when the decision maker makes a negative decision despite the person having the requisite qualifications. The decision maker in this case would make the opposite decision if the person belonged to the majority class. The decision maker, whether conscious or subconscious, is acting in a biased manner against the minority class where the probability of a favorable outcome for the minority group is less than that of the majority group. Expressed mathematically this is:

\[P(y=1|p=minority) < P(y=1|p=majority)\]

Thus, ground truth labels in certain applications become biased. Measuring the performance of a machine learning model against biased labels without any bias mitigation poses problems of perpetuating bias downstream. 

The first goal of \algo is to debias ground truth labels. This is an inherently ill-posed problem because (1) the real unbiased ground truth is not known and (2) asking human experts to relabel is either infeasible (e.g., prohibitive cost and insufficient data to make decisions) or not advisable (i.e., biases may still be present) or both. The recommended approach in the literature is Situation Assessment (SA). SA can correct the labels of both majority and minority groups. We propose a variant of this based on the fundamental observation that we first want to correct labels only of the minority group in only one direction (0 $\rightarrow$ 1). This is inline with the real-world biases acting against minority group. This process also ensures the positive biases received by the majority group are reflected in the correction for minority group. We call this process \textsc{FairMin} because the minority group's data is corrected. The process is as follows:

\begin{enumerate}
\item \textbf{Split data into majority and minority groups:} Create training and validation data consisting only of the majority group.
\item \textbf{Train a classifier on majority group:} Build a model using only the majority group's data. This can be an ensemble of classifiers or any single classification algorithm. This ensures the model only learns the patterns from the majority group.
\item \textbf{Run inference on minority group:} Get predictions for minority class using the trained model in the previous step. 

\item \textbf{Flip labels of minority group:} For a given data point where the ground truth label is 0 and the model predicts 1, flip the label, i.e. the label transforms ‘0 $\rightarrow$ 1’ for all instances where prediction=1. We can can use a hold-out set to determine threshold for ‘0 $\rightarrow$ 1’ label change so that DIR is close to 1.

\item \textbf{Concatenate majority and minority datasets:} After flipping labels in the minority dataset, combine the majority dataset with the modified/debiased minority dataset. The resulting dataset can be considered to be debiased. 
\end{enumerate}

In several scenarios, not only does the bias manifest against the minority group, the data also has bias in favor of the majority group. To remove such bias, we complement the above process and debias the majority group dataset. Similar to \textsc{FairMin}, we split the dataset by majority/minority groups, but train a classifier only on the minority dataset and flip the labels of the majority group from 1 to 0 if the prediction=0. We call this process \textsc{FairMaj} and the detailed steps are as follows:

\begin{enumerate}
\item \textbf{Split data into majority and minority groups:} Create training and validation data consisting only of the minority group.
\item \textbf{Train a classifier on minority group:} Build a model using only the minority group's data. This can be an ensemble of classifiers or any single classification algorithm. This ensures the model only learns the patterns from the minority group.
\item \textbf{Run inference on majority group:} Get predictions for majority class using the trained model in the previous step. 

\item \textbf{Flip labels of majority group:} For a given data point where the ground truth label is 1 and the model predicts 0, flip the label, i.e. the label transforms ‘1 $\rightarrow$ 0’ for all instances where prediction=0. We can can use a hold-out set to determine threshold for ‘1 $\rightarrow$ 0’ label change so that DIR is close to 1.

\item \textbf{Concatenate majority and minority datasets:} After flipping labels in the majority dataset, combine the minority dataset with the modified/debiased majority dataset. The resulting dataset can be considered to be debiased. 
\end{enumerate}

For \algo, we first run \textsc{FairMin} and then, optionally, run \textsc{FairMaj}. Depending on the level of bias inherent in the data, whether it is present in the minority decisions or majority decisions, we can choose to run \textsc{FairMin} or \textsc{FairMaj}, or both.

\section{Synthetic Dataset for Classification} \label{syngen:sec}
We validate the label flipping using synthetically generated data. By injecting bias into synthetic data, we have the ability to track where the bias was added, an attribute of the synthetic data that is not possible in real-world datasets. Consider a classification problem which has some protected attributes like gender and race. To generate biased synthetic data for this, we need the following.

\begin{itemize}
\item \textbf{Independent Variables:} These are the aspects of data such as numerical, categorical, and textual features. The variables necessarily include protected attributes
\item \textbf{Data Model:} This is the underlying data generation process. The data generation depends on all the independent variables excluding protected attributes. Inclusion of protected attributes is optional. We can use a range of techniques like linear regression, mixture models, decision trees, etc. to generate clean and unbiased data.
\item \textbf{Noise:} Real-world data is noisy. Both features and labels can be noisy. We use a noise model to introduce noise in non-protected attributes and run the model to generate noisy labels. If the data model is simple, this may not introduce noise in labels. So also introduce noise in labels. \textit{The noise is independent of protected attributes.}
\item \textbf{Bias:} \textit{Bias can be considered systematic label noise which depends on protected attributes.} The noise is systematic in the sense that it is unidirectional: either 0$\rightarrow$1 (for majority) or 1$\rightarrow$0 (for minority) based on protected attributes. The severity of bias is given by bias probability.
\end{itemize}

\subsection{Synthetic Data Generation}
\begin{figure*}
\begin{center}
\includegraphics[scale=0.5]{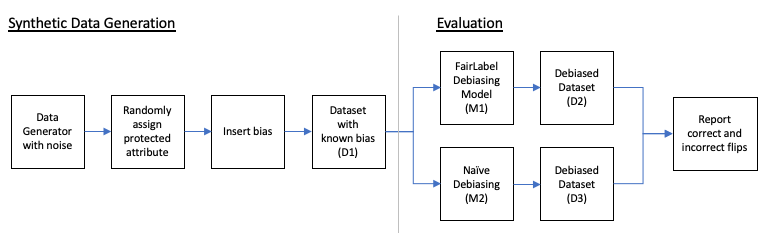}
\end{center}
\caption{\label{synth-eval:fig}Synthetic data generation and validation of \algo}
\end{figure*}

Real-world datasets have both bias and noise. Noise can affect both features and labels. \textit{Noise is independent of the protected class.} We quantify noise as $\epsilon$. Bias can be considered noise except that \textit{bias is based on the protected attribute and is unidirectional}. See Figure~\ref{synth-eval:fig} for Synthetic data generation as well as its use in \algo\ evaluation.

\begin{figure*}
\begin{center}
\includegraphics[scale=0.5]{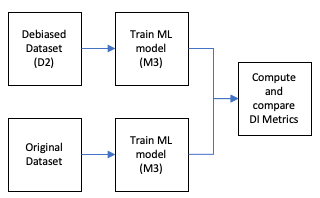}
\end{center}
\caption{\label{di-eval:fig}Evaluation of DI metrics}
\end{figure*}

\subsection{Linear Synthetic Data Generation Example}
\label{sec:synthetic}
\textbf{Clean Data:} As an example, let us consider 10 unprotected numerical features ($x$) and 1 protected categorical feature ($z$) for a linear classification problem.
\[ y = f(a^T x + b) \]
where $a$ are the coefficients, $b$ is the intercept, and $f(\cdot)$ is the logistic function. The model is specified by 11 random coefficients for $x$ and $b$. We now randomly generate $N$ data points based on random $x$. The protected attribute $z$ is also randomly generated (with certain distribution) but not fed into the model.

\textbf{Noisy Data:} We now introduce noise to the data.
\[ y' = f(a^T x' + b + \epsilon) \]
where $x'$ and $\epsilon$ are noisy versions.

\textbf{Biased Data:} We now fix the protected attribute value ($v$), bias direction ($d$) and bias severity ($p$).
\begin{enumerate}
\item Loop over records with protected attribute value, $z = v$.
\begin{enumerate}
     \item Choose a record with probability $p$.
     \item Change the label in the direction $d$.
\end{enumerate}
\item Store the changes as metadata.
\end{enumerate}

\section{Metrics for Debiasing Task} \label{metrics:sec}
Assume that the ground truth is known. Assume negative bias based on the unprotected attribute. In this case, \algo\ recommends $0 \rightarrow 1$ flips for the minority class. Some of the flips are correct and TPR is the fraction of correct flips. TNR is the fraction of missed flips. FPR and TNR are not applicable. We rename these metrics as Correct Flip Rate (CFR) and Missed Flip Rate (MFR). We report CFR and MFR for the majority class too. In case of the majority class, the metrics are reported for $1 \rightarrow 0$ flips. 

It is not possible to measure CFR and MFR in real-world datasets. Instead, we consider standard observational metrics related to bias and fairness. Definitions are as follows:

\textbf{Demographic parity or statistical parity:} it suggests that a predictor is unbiased if the prediction \(\hat{y}\) is independent of the protected attribute p so that:
\[Pr(\hat{y}|p)=Pr(\hat{y})\]

\textbf{Disparate Impact Ratio (DIR):} The ratio of the demographic parity: 
\[\frac{ Pr(\hat{y}=1|p=minority)}{Pr(\hat{y}=1|p=majority)}\]

\textbf{Disparate Impact Difference (DID):} The difference in demographic parity: \[Pr(\hat{y}=1|p=minority)-Pr(\hat{y}=1|p=majority)\]

\section{Experimental Results} \label{expt:sec}

\subsection{Synthetic Dataset}
We ran several experiments comparing \algo\ to a Naive ML model for debasing using three different synthetic datasets listed below. The baseline Naive approach is described in Section~\ref{baseline:sec}.
\begin{itemize}
    \item \textbf{Linear:} Linear dataset created by logistic function as described in section \ref{sec:synthetic}.
    \item \textbf{Clusters around n-hypercubes:} This initially creates clusters of points normally distributed (std=1) about vertices of an n-informative-dimensional hypercube and assigns an equal number of clusters to each class. It introduces interdependence between these features and adds various types of further noise to the data. We use 8-dimensional hypercube with an edge length of 0.5.
    \item \textbf{Gaussian Quantiles:} This classification dataset is constructed by taking a multi-dimensional standard normal distribution and defining classes separated by nested concentric multi-dimensional spheres such that roughly equal numbers of samples are in each class (quantiles of the distribution).
\end{itemize}
Additional details of the data generation are listed in Appendix~\ref{datagen:app}.
For each of the synthetic datasets, we generated 100,000 samples with 10 features and a binary class for the label. In addition, we experimented with Logistic Regression, Random Forest, and Gradient Boosted Tree algorithms for the underlying ML ensemble model, with an 80/20 random split between train and test sets. We vary the bias injection rate in the synthetic data to understand how the amount of bias inherent in the data affects \algo.

\subsection{Baseline for Debiasing} \label{baseline:sec}
The baseline for the debiasing does not take directionality of bias into consideration. The debiasing is based on training an ML model on the full data and flipping labels of the minority class from 0 to 1 based on classifier prediction. We call this the Naive approach. Figure~\ref{synth-eval:fig} shows how we perform the baselining.

\subsection{Benchmark Datasets}
We use three benchmark datasets: UCI Adult, German Credit Risk, and Compas.

\begin{itemize}
 \item \textbf{UCI Adult:} The dataset has 14 attributes and task is to predict whether income exceeds \$50K/yr based on census data. Also known as "Census Income" dataset.The protected attributes in the dataset are age, gender, and race. We use gender as the protected attribute.
 \item \textbf{German Credit Risk:} The dataset contains 20 attributes and the task is to classify people described by a set of attributes as good or bad credit risks. The protected attribute is sex.
 \item \textbf{Compas:} The dataset has 10 attributes and the target is prediction of two-year recidivism. The protected attributes are race and sex. For our purpose, we used race binarized to ‘Caucasian’ (majority) and ‘African-American’ (minority).
\end{itemize}
Table~\ref{data:tab} summarizes the data sets.

\begin{table*}
\caption{\label{data:tab} Summary of datasets.}
\begin{center}
\begin{tabular}{|c|r|r|c|c|}
\hline
Dataset & Number of records & Number of features & Target & Protected Attribute \\
\hline
UCI Adult & 48,842 & 14 & salary$>$\$50k/yr & gender  \\
German Credit Risk & 1,000 & 20 & credit risk  & gender  \\
Compas & 6,167 & 10  & two-year recidivism  & race  \\
\hline
\end{tabular}
\end{center}
\end{table*}

\begin{table*}[!htb]
\centering
\caption{\label{results:tab}\algo\ vs Naive model overall results, bias injection rate = 0.2}

\begin{tabular}{l r r r}
\hline
 \textbf{Model} & \textbf{Correct Flip Rate (CFR)} & \textbf{Missed Flip Rate (MFR)}& \textbf{F1-score} \\
\hline
Naive model & 0.7192 $\pm$ 0.1647 & \textbf{0.1145 $\pm$ 0.0640} & 0.7800 $\pm$ 0.09521 \\
\algo\ & \textbf{0.8668 $\pm$ 0.0727} & 0.2002 $\pm$ 0.1040 & \textbf{0.8818 $\pm$ 0.0364} \\
\hline
\textit{Gain} & 0.1476 & -0.0858 & 0.1017  \\
\hline
\end{tabular}
\end{table*}

\begin{table*}[!htb]
\caption{\label{di:tab}Disparate impact ratio (DIR) of \textit{aif360} datasets}
\centering

\begin{tabular}{l r r r}
\hline
\textbf{Dataset} & \textbf{UCI} & \textbf{German Credit Risk} & \textbf{Compas} \\
\hline
Original (not debiased) & 0.3122 & 0.896 & 0.618 \\
Debiased (\algo) & 0.665 & 1.03 & 1.16 \\
\hline
\textit{Gain} & 0.356 & 0.134 & 0.542 \\
\hline
\end{tabular}

\end{table*}

\subsection{Results}

\subsubsection{Results on Synthetic Datasets}
We compare the Correct Flip Rate (CFR), Missed Flip Rate (MFR) of datasets D2 and D3, and F1-score (of Models M1 and M2) across multiple synthetic datasets where the debiased datasets produced by \algo\ and the Naive models (D2 and D3) are compared against the dataset with known bias (D1). Table \ref{results:tab} summarizes the results of \algo\ across these metrics with the bias injection rate held constant at 0.2. Results show that \algo\ corrects more bias than the Naive model. Also, \algo\ does not sacrifice accuracy in order to capture more bias flips as it outperforms the Naive model on F1-score. This is true across the entire range of synthetic datasets. Bias acts as noise to an ML model and \algo\ is not adversely affected by bias in the minority class because the \algo\ is only trained on the majority class.

\algo\ incorrectly flips labels at a slightly higher rate than the Naive approach. This pattern holds true across datasets and model types. i.e there is a penalty for flipping labels in a unidirectional manner. In real world-applications, this means some people in the minority class will be "incorrectly" boosted.

\begin{figure*}
\begin{center}
\includegraphics[scale=0.5]{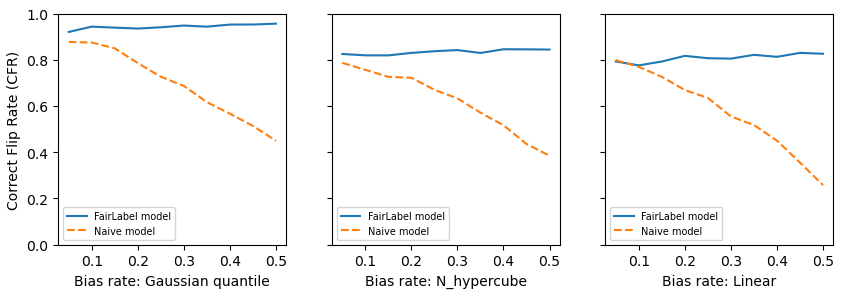}
\end{center}
\caption{\label{cfr_v_bias:fig}CFR vs bias injection rate}
\end{figure*}

\begin{figure*}
\begin{center}
\includegraphics[scale=0.5]{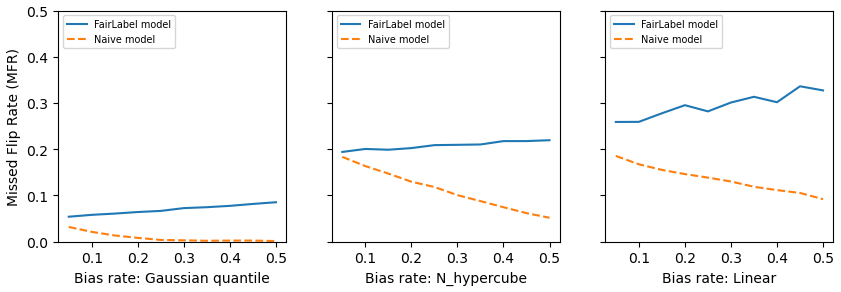}
\end{center}
\caption{\label{mfr_v_bias_min:fig}MFR vs bias injection rate}
\end{figure*}

\begin{figure*}
\begin{center}
\includegraphics[scale=0.5]{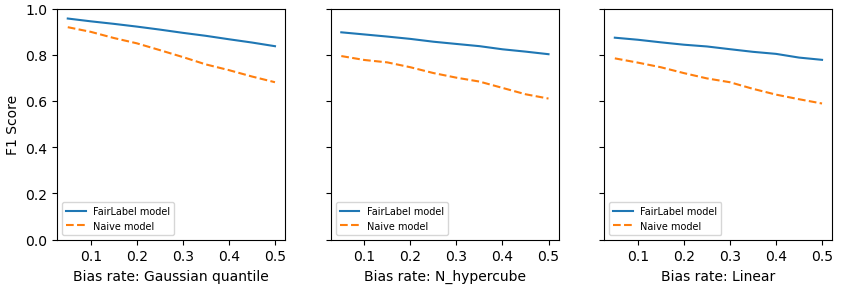}
\end{center}
\caption{\label{f1score_v_bias:fig}F1-score vs bias injection rate}
\end{figure*}

In general, \algo\ is more robust to the amount of bias present in the data in terms of CFR and F1-score vs. the Naive models across different levels of bias. For the data in Figures \ref{cfr_v_bias:fig}, \ref{mfr_v_bias_min:fig}, and \ref{f1score_v_bias:fig}, we add bias to the synthetic datasets at various proportions to see how the bias proportion affects the CFR, MFR, and F1-score, respectively.

\begin{figure*}
\begin{center}
\includegraphics[scale=0.5]{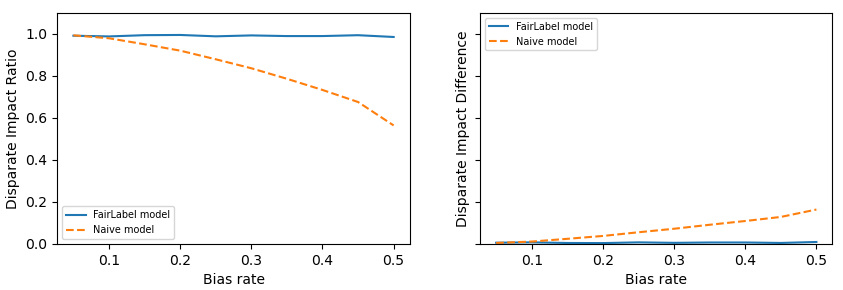}
\end{center}
\caption{\label{di_v_bias:fig}Disparate Impact vs bias injection rate, XGboost}
\end{figure*}

The US Equal Employment Opportunity Commission states that bias is acceptable when DIR is between 0.8 and 1.0. Checking DIR in our models trained on synthetic datasets, we see that models trained on debiased datasets produced by \algo\ does better than those trained on the original (biased) datasets, regardless of the amount of bias in the data, with DIR close to 1 and DID difference to be close to 0. See Figure \ref{di_v_bias:fig}.

\subsubsection{Results on Benchmark Datasets}
The advantage or running analyses on synthetic datasets is the ability to measure CFR and MFR because we know with certainty which labels are biased. Real-world datasets, however, do not have this advantage. Instead we look at aggregate-level metrics, such as as demographic parity and disparate impact. 

We ran \algo\ on the datasets Adult, German Credit Risk, and Compass. Figure~\ref{di-eval:fig} shows the evaluation setup. We then ran the AI Fairness Package (\href{https://github.com/Trusted-AI/AIF360}{aif360}) and checked Disparate Impact Ratio both on models trained on the debiased dataset created by \algo\ and on models trained on the original datasets. The results are in Table~\ref{di:tab}. Across each dataset, the DIR improved from +0.356 (UCI) to +0.542 (Compas), showing that \algo\ has the ability to reduce disparity between groups.

\section{Conclusions} \label{conc:sec}
In this paper, we have considered the fundamental issue of bias in ground truth labels and have proposed an intuitive approach to address the issue: bias is directional and affects subsets of data differently. The algorithm to codify the intuition has to parts: \textsc{FairMin} and \textsc{FairMaj}. \textsc{FairMin} addresses the problem of negative bias for minority attribute while \textsc{FairMaj} addresses the potential positive bias for majority attribute. The final algorithm, \algo, applies both \textsc{FairMin} and \textsc{FairMaj} iteratively to remove bias. We have defined metrics to characterize the performance of the algorithms and have proposed a synthetic data generation framework for validating the approaches. We show that \algo\ reduces DIR by almost 55\% in some cases. We believe the results presented in the paper will inspire the adoption of \algo\ to other fairness problems. We also hope that the synthetic data generation will stimulate other data generation as well as label debiasing approaches because it makes benchmarking feasible. Our own goal is to extend \algo\ to other modalities like text and images.

\bibliographystyle{unsrt}
\bibliography{bias.bib}

\clearpage
\appendices

\textbf{\Large Appendix}
\vspace*{2em}

\begin{table*}[ht]
 \centering
 \caption{Metrics used as fairness criteria}
\begin{tabular}{|c|p{15em}|c|} 
\hline
Metric & & Definition \\
\hline
EO & Equalized Odds &  P(\^{Y}=1$|$A=0,Y =y) = P(\^{Y}=1$|$A=1,Y =y) , y$\in$\{0,1\} \\ \hline
DP & Demographic Parity & P(\^{Y} $|$A = 0) = P(\^{Y}$|$A = 1) \\ \hline
DI & Disparate Impact &  $\frac{\frac{T P_{p}+F P_{p}}{N_{p}}}{\frac{T P_{u}+F P_{u}}{N_{u}}}$ \\ \hline
EOO & Equal of Opportunity & $\frac{TP_{p}}{TP_{p}+FN_{p}}-\frac{TP_{u}}{TP_{u}+FN_{u}}$ \\ \hline
KNNC & K-Nearest Neighbors Consistency & \\ \hline

ABAD & Absolute Balanced Accuracy Difference & $\left | \frac{1}{2} \left [ TPR_{p} + TNR_{p}  \right ] - \left [ TPR_{u} + TNR_{u}  \right ] \right |$ \\ \hline
AAOD & Absolute Average Odds Difference & $\left | \frac{\left ( FPR_{u} + FNR_{p}  \right ) - \left ( TPR_{u} + TPR_{p}  \right )}{2} \right |$ \\ \hline
AEORD & Absolute Equal Opportunity Rate Difference &  $\left | TPR_{p} - TPR_{u} \right |$ \\ \hline
SPD & Statistical Parity Difference & $ \frac{TP_{p} +FP_{p}}{N_{p}}-\frac{TP_{u}+FP_{u}}{N_{u}}$ \\ 

\hline

\end{tabular}

\label{tab:metrics_fairness}
\end{table*}

\section{Fairness Measures} \label{metrics:app}
Table \ref{tab:metrics_fairness} summarizes standard fairness metrics.

\section{Details of Synthetic Data Generation} \label{datagen:app}
 N\_hypercubes and Gaussian quantiles are from the \verb|sklearn| package and the linear dataset was a custom function shown in \ref{code:app}



\clearpage
\onecolumn
\section{Linear Synthetic Data Generation Code}\label{code:app}

\begin{verbatim}
def generate_linear_dataset(n_samples, n_features, p_noise, seed):
    import pandas as pd
    
    #compute sample counts
    n_samples_perfect = int(n_samples*(1-p_noise))
    n_samples_noise = n_samples - n_samples_perfect

    #weights of the model
    w = generate_random_coefficients(n_features,seed=seed)
    
    #random X's
    X = generate_random_x(n_features,n_samples_perfect,seed=seed+20)
    b = 0
    
    #compute y (perfect y)
    probs = sigmoid(np.dot(X,w) + b)
    y = np.array([1 if i > 0.5 else 0 for i in probs]).reshape(n_samples_perfect,)
    
    #noisy data
    X_noise = generate_random_x(n_features,n_samples_noise,seed=seed+40)
    y_noise = generate_random_binary(n_samples_noise, seed=50)

    #combine perfect data with noisy data
    X_full = np.concatenate((X,X_noise))
    y_full = np.concatenate((y,y_noise))
    
    return X_full, y_full
    
\end{verbatim}

\end{document}